# On Transformations between Probability and Spohnian Disbelief Functions

Phan H. Giang and Prakash P. Shenoy

University of Kansas School of Business, Summerfield Hall
Lawrence, KS 66045-2003, USA
pgiang@ukans.edu, pshenoy@ukans.edu

## Abstract

In this paper, we analyze the relationship between probability and Spohn's theory for representation of uncertain beliefs. Using the intuitive idea that the more probable a proposition is, the more believable it is, we study transformations from probability to Spohnian disbelief and vice-versa. The transformations described in this paper are different from those described in the literature. In particular, the former satisfies the principles of ordinal congruence while the latter does not. Such transformations between probability and Spohn's calculi can contribute to (1) a clarification of the semantics of non-probabilistic degree of uncertain belief, and (2) to a construction of a decision theory for such calculi. In practice, the transformations will allow a meaningful combination of more than one calculus in different stages of using an expert system such as knowledge acquisition, inference, and interpretation of results.

## 1 Introduction

In [19, 20] Spohn describes a non-probabilistic theory for epistemic belief representation. One notable advantage of this theory compared to probability theory is in representing the notion of plain belief that is supposed to be deductively closed. The deficiency of probability theory to do such a job has been demonstrated by the well-known Lottery Paradox that describes the situation in which each of a million lottery players has no practical chance to win the jackpot, nevertheless, one among them will surely win. Another nice feature of Spohn's theory is its naturally defined concept of conditionals which is similar to its probabilistic counterpart in many aspects. Hunter [9] and Shenoy [13] exploit that structural similarity to implement Spohn's theory using computational architectures traditionally used for probability such as valuation networks and Bayesian networks. From a theoretical perspective, on one hand, Dubois and Prade [4] observe that the basic representation in Spohn's theory, the disbelief function, can be interpreted as the negative of the logarithm of a possibility function. On the other hand, interpretation of disbelief values as infinitesimal probabilities makes Spohn's theory tightly related to Adams' study of $\epsilon$-semantics for default reasoning [1]. Thus, Spohn's theory is well positioned in the web of quantitative approaches to represent and reason about uncertain beliefs. In the remainder of this section, we shall briefly review the theory and ask questions that motivate the study of Spohn's belief-probability transformations.

Let $\Omega$ denote a set of possible worlds. For simplicity, we assume $\Omega$ is finite, $|\Omega| = n$. We use $\omega$ (perhaps with subscripts) to denote a world, i.e. $\omega \in \Omega$.

A Spohnian disbelief function $\delta$ is defined as a mapping

$$\delta : \Omega \to Z^+$$

where $Z^+$ is set of non-negative integers. $\delta$ satisfies the following axiom:

S1 $\quad \min_{\omega \in \Omega} \delta(\omega) = 0.$

An extension of $\delta$ to the set of all nonempty subsets of $\Omega$ is defined as follows

S2 $\quad \delta(A) = \min_{\omega \in A} \delta(\omega)$ for all $A \subseteq \Omega$.

For $A \subseteq \Omega$ and $\omega \in A$, the conditional disbelief function $\delta(\omega|A)$ is defined as

S3 $\quad \delta(\omega|A) = \delta(\omega) - \delta(A).$

It is easy to verify that $\delta(\omega|A)$ is a disbelief function on the (contracted) state space $A$ i.e., it satisfies $S1$. Therefore, for any subset of $A$, axiom $S2$ can be applied to determine the disbelief value. Analogous to probability theory, the conditional degree of disbelief for a set $B$ satisfying $B \cap A \neq \emptyset$ is defined as

S4  $\delta(B|A) = \min\limits_{\omega \in A \cap B} \delta(\omega|A)$.

Given disbelief function $\delta$, a Spohnian belief function $\beta : 2^\Omega \to Z$ (where $Z$ is the set of whole numbers) can be defined as follows

$$\beta(A) = \begin{cases} -\delta(A) & \text{if } \delta(A) > 0 \\ \delta(\neg A) & \text{otherwise.} \end{cases}$$

If one defines "proposition $A$ is *believed* with degree $m(> 0)$" to mean $\beta(A) = m$, then it is easy to show that the set of propositions believed with respect to a disbelief function is deductively closed. This desirable property supports the idea of using disbelief functions to represent epistemic beliefs.

To see the similarity between Spohnian disbelief functions and probability functions, let us list the probabilistic counterparts for the axioms S1 through S4. A probability function on the set $\Omega$ is a mapping

$$p : \Omega \to [0, 1]$$

that satisfies following axioms.

P1  $\sum_{\omega \in \Omega} p(\omega) = 1$.

An extension of the function $p$ to the set of all subsets of $\Omega$ is defined as

P2  $p(A) = \sum_{\omega \in A} p(\omega)$ for all $A \subseteq \Omega$.

The conditional probability function given $A$ is defined for $\omega \in A$ as

P3  $p(\omega|A) = \frac{p(\omega)}{p(A)}$

and for $B \subseteq \Omega$

P4  $p(B|A) = \sum_{\omega \in B} p(\omega|B)$.

The pairwise similarity of $Si$ and $Pi$ is obvious. While a qualification $\omega \in A$ is used in the definition of conditional disbelief function, for probability such qualification seems unnecessary (for $\omega \notin A$, $p(\omega|A) = 0$). However, if we follow the convention $\delta(\omega|A) = \infty$ for $\omega \notin A$, then even this difference disappears.

Until now, we are satisfied with the suggestion that the value assigned to a proposition by a disbelief function somehow reflects the firmness or strength of (subjective) belief in that proposition. Although Spohn in [19, 20] used the term "ordinal conditional function" instead of "disbelief function" (introduced by Shenoy [13]), he endorsed the usage of the latter term that bore an obvious intuitive semantic. For the task of constructing a philosophical theory about epistemic beliefs that Spohn clearly engaged in, such an informal interpretation of disbelief functions is obviously helpful for exposition but not required as an absolute necessity. Abstract constructs are often sufficient for that kind of purpose. However, for those who want to build (computer) applications, such interpretations are often too abstract to be useful. They have to answer questions such as how to extract a disbelief function from available evidence, data, and human expertise; how to justify the use of one uncertainty calculus over another (for representation of epistemic states) in a specific situation; how to interpret the results or how to make use of results in decision problems. We believe that these concerns of practitioners could be addressed by a further exploration on the relationship between Spohnian disbelief and probability functions.

The first hint of such a relationship has originated from Spohn. He suggests that "$A$ is disbelieved with degree $i$" ($\delta(A) = i$) is equivalent to "$p(A)$ is of the same order as $\epsilon^i$" for some probability function $p$ and any infinitesimal $\epsilon$. For the case of uncountable ordinals originally considered in [19, 20], this interpretation offers an explanation for the minimization operation in the definition of disbelief value for a non-atomic proposition and the substraction operation in the definition of conditionals. But for a practical situation of a finite state space as we assume here, the condition about the order comparison based on infinitesimals is not very informative even if infinitesimals can be operationalized as close to zero. This approximation, in fact, has been used in [2, 8] by Darwiche and Goldszmidt, and Henrion et al.

The operationalization of infinitesimals by close-to-zero numbers creates another problem. That is, such transformations may lead to counter-intuitive consequences. The intuition that is at risk is the monotonicity of disbelief values relative to probability because a proposition of lower degree of disbelief is intuitively conceived as having a higher probability. To clarify the point, let us look at the experimental results in [2, 8]. Using $\epsilon$–rule: "if $\epsilon^{k+1} < p(A) \leq \epsilon^k$ then $\delta(A) = k$", the authors apply Spohn's theory and probability calculus with various close-to-zero values of $\epsilon$ and for many instances of a car troubleshooting problem and then compare the fault orderings resulting from these applications. Because of the mentioned intuition, it is desirable that the orderings of possible faults according to (descending) probabilities produced by probabilistic calculation and (ascending) disbelief degrees produced by Spohn's calculation are the same. However, the results of experiments show that those orderings do not always coincide. For an illustration, let us use the following simple example.

**Example:** Suppose we have $\Omega = \{\omega_1, \omega_2, \omega_3, \omega_4\}$ with a given probability distribution. Using $\epsilon$–rule, we have the following table.

| $\omega$ | $p$ | $\delta_{\epsilon=.2}$ |
|---|---|---|
| $\omega_1$ | 0.5185 | 0 |
| $\omega_2$ | 0.2308 | 0 |
| $\omega_3$ | 0.1538 | 1 |
| $\omega_4$ | 0.0969 | 1 |

Now let $A = \{\omega_2\}$ and $B = \{\omega_3, \omega_4\}$, we have $p(A) < p(B)$ or "$A$ is less probable than $B$" but after transformation with $\epsilon = 0.2$ we have $\delta(A) = 0 < \delta(B) = 1$ or "$A$ is less disbelieved than $B$". The main goal of this work is to find a remedy to this problem.

Earlier in this section, we cited Hunter's and Shenoy's works showing that automated inference with disbelief functions can be easily implemented using architectures developed for probabilistic inference, in fact, with simpler computation [7]. For example, Shenoy's valuation-based system framework can be used for probabilistic as well as non-probabilistic calculi such as possibility theory and Spohnian disbelief functions [13, 14, 15]. More striking is the fact that the axioms that allow local computation in valuations-based systems [17] are satisfied by possibility theory and Spohn's epistemic-belief theory. This observation allows one to develop a expert system shell that offers à-la-carte calculi for reasoning under uncertainty [16]. But a problem remains. Although the system has a common computational engine for different calculi, it is still not a genuine combination of these calculi. The system requires a user to pre-select a calculus and then lets the user construct a knowledge base and interpret the results within that calculus. So, a transformation between probabilistic and non-probabilistic calculus that will enable a dynamic exchange inside the system is highly desirable. For example, using Spohn's notion of plain beliefs we can encode opinions expressed by experts who may feel reluctant to commit to exact numerical probabilities required by Bayesian networks. Then, we can combine that kind of information with statistical data using probability, make inferences with the combined data, and interpret the results in the calculus most convenient to users.

## 2   From Probability to Spohnian Disbelief

In this section, we shall consider the problem of finding transformations from probability functions to Spohnian disbelief functions. Using the relationship pointed out by Dubois and Prade [4] that a disbelief function can be interpreted as a the negative of the logarithm of a possibility function, we can use this mapping to go from probability to possibility and vice-versa.

Denote by $\mathcal{P}$ the set of probability distributions over set $\Omega$ and $\Delta$ the set of Spohn's disbelief functions over the same set. We consider the transformation

$$T : \mathcal{P} \to \Delta \qquad (1)$$

As discussed in the previous section, it is difficult to explain the exact semantics of Spohnian disbeliefs. For example, how can one interpret the statement "a proposition $A$ is disbelieved to degree $n$". But we do know that the values of a disbelief function are used to rank propositions. So, it is reasonable to suggest the following principle.

**Definition 1 (Principle of ordinal congruence I)** *Transformation $T$ is said to satisfy the principle of ordinal congruence if $\forall p \in \mathcal{P}$ and $\forall A, B \subseteq \Omega$, if $p(A) \geq p(B)$ then $T(p)(A) \leq T(p)(B)$.*

In plain language, the principle of ordinal congruence says that the more probable a proposition is, the less disbelievable it should be. This principle is similar to that used by Dubois et al. and Delgado and Moral in considering consistent possibility to probability functions [3, 6].

Note that the set of congruent transformations is not empty. A trivial transformation that matches every probability distribution in $\mathcal{P}$ to the vacuous disbelief function $o$ is obviously a congruent transformation. The vacuous disbelief function $o$ on $\Omega$ is defined as $\forall \omega \in \Omega, o(\omega) = 0$.

Since the cardinality of $[0, 1]$ - the range of a probability function - is uncountable, $|\mathcal{P}|$ is also uncountable. Similarly, we know that $|\Delta|$ is countable because $|N|$ is countable. So, the transformation $T$ is many-to-one.

Comparing a probability distribution and its associated Spohn's disbelief function, notice that $T$ is a "coarsening" process. For probability, it is possible that each subset of $\Omega$ has a distinct probability i.e. $2^n$ subsets have $2^n$ different values. As a simple example, consider the probability distribution $p$ such that $p(\omega_i) = 2^{-i}.Z$ where $i$ is the subscript of $\omega_i$, an arbitrary labeling of the elements of $\Omega$, and $Z$ is the normalization constant. It is not difficult to show that for $A, B \subseteq \Omega$ if $A \neq B$ then $p(A) \neq p(B)$. Formally, if we define $C_x = \{A \subseteq \Omega | p(A) = x\}$, then $|\{x | C_x \neq \emptyset\}| \geq n$. We can show that the equality happens if and only if $p(\omega) = \frac{1}{n}$ for all $\omega \in \Omega$. In contrast, because of the minimization operation, it is evident that the number of levels a Spohn's disbelief function $\delta$ has on the set of subsets of $\Omega$ is less or equal to $n$. The equality happens only if the degrees of disbelief for singletons are all distinct. Therefore, a transformation from probability function to disbelief function will match each level of the latter to one or more levels of the former.

Intuitively, the larger the number of levels an uncertainty measure has, the more informative it is. Obviously, $o$ is the most coarse disbelief function. We formalize this intuition by a definition. Denote by $|T(p)|$ the number of levels $T(p)$ has on $2^\Omega$ i.e. $|\{x|\exists \omega \in \Omega, T(p)(\omega) = x\}|$.

**Definition 2** *Let $T_1$ and $T_2$ be two transformations. We say $T_2$ is coarser than $T_1$ if $|T_1(p)| \geq |T_2(p)|$ for all $p \in \mathcal{P}$*

Because we want to use a measure of uncertain belief to differentiate among propositions we shall set our goal to find a least-coarse congruent transformation. We need the following definition to give an upper bound for the number of levels produced by a congruent transformation.

**Definition 3 (Leap indices)** *For a non-increasing sequence $Q$ of numbers $q_1 \geq q_2 \geq \ldots \geq q_n$, the leap index set $L_Q$ is defined as $L_Q = \{i | q_i > \sum_{j=i+1}^{n} q_j\}$.*

In this definition, $n$, the index of smallest element of $Q$, is not a leap index. Informally, an index will be included in the leap set if the "mass" associated with that index is (strictly) greater than sum of those associated with all following indices. In general, from a set of numbers $\{q_i\}$ we can arrange them in more than one non-increasing sequences. Therefore, we can have more than one leap index set. For example, if $q_i = q_j$ then $q_i$ and $q_j$ can swap their positions in a non-increasing sequence to obtain another non-increasing sequence. However, it is easy to show that the cardinality of $L_Q$ is independent of such rearrangement, and therefore, it is a characteristic for the set $\{q_i\}$. We also note that the cardinality of a leap index set can range from 0 (for example in case $q_i = q_j$ for different $i, j$) to $|\{q_i\}| - 1$.

**Lemma 1** *If $T$ is a congruent transformation then $|T(p)| \leq |L_p| + 1$.*

**Proof:** Let $|L_p| = m$. Suppose to the contrary that $|T(p)| > m + 1$. Let $\Omega_i = \{\omega | T(p)(\omega) = i\}$. For $i = 0, 1, \ldots, m, m+1$ we have $|\Omega_i| = k_i > 0$. For $w \in \Omega_i$ and $v \in \Omega_j$, $i < j$, because $T(p)$ is congruent, we have $p(w) > p(v)$. So we can, first, to locally rearrange each $\Omega_i$ in non-increasing order according to probability. Then, concatenate $m + 2$ locally non-increasing sequences into one non-increasing sequence. In other words, we can have $p_1 \geq p_2 \geq \ldots \geq p_n$ such that $\{\omega_i | i = 1, \ldots k_0\} = \Omega_0$, $\{\omega_i | i = k_0+1, \ldots k_0+k_1\} = \Omega_1$ and so on, and $k_0 + k_1 + \ldots + k_{m+1} = n$. For each $0 \leq i \leq m + 1$, define $G_i = \{\omega_j | j > k_0 + k_1 + \ldots k_i\}$. In other words, $G_i$ represents the union of the sets $\Omega_j$ for $j > i$. On one hand, by definition, we have $T(p)(G_i) = i + 1$. On the other hand, $T(p)(\omega_{s_i}) = i$ where $s_i = k_0 + k_1 + \ldots k_i$. By congruence of $T(p)$, we infer that

$$p(\omega_{s_i}) > p(G_i) = \sum_{j > s_i} p(\omega_j).$$

That means the set $\{s_i | i = 0, \ldots, m\}$ is a subset of a leap index set. Obviously, $|\{s_i | i = 0, \ldots, m\}| = m + 1$, but we have assumed $|L_p| = m$. That contradicts the hypothesis $|T(p)| > m + 1$. ∎

For a given probability distribution, we re-label the elements $\omega$ such that $p(\omega_1) \geq p(\omega_2) \geq \ldots \geq p(\omega_n)$. Let $p_i$ denote $p(\omega_i)$.

**Definition 4 (Function $T$)**

Input: A sequence of probabilities $(p_1, p_2, \ldots, p_n)$.

Output: A sequence of disbelief degrees $(d_1, d_2, \ldots, d_n)$.

$r = 0$         % $r$ is disbelief counter, initially equal 0.

$M = 1$         % $M$ is remaining mass, initially equal 1.

for $i = 1$ to $n$
    $d_i = r$         % $d_i$ is disbelief degree of $\omega_i$.
    $M = M - p_i$     % $p_i$ is probability of $\omega_i$.
    if $p_i > M$ then $r = r + 1$
end

In other words, this simple algorithm runs once through $\Omega$ in the descending order of probabilities. Initially, the most probable world $(\omega_1)$ gets disbelief degree 0. Disbelief degree counter increases by one at each leap index. We have the following theorem.

**Theorem 1** *The function $T$ in Definition 4 is a least-coarse congruent probability to Spohnian disbelief transformation function.*

**Proof:**

*Congruence.* Let $A$ and $B$ be singletons i.e. $A = \{\omega_i\}$ and $B = \{\omega_j\}$. If $p(\omega_i) > p(\omega_j)$, then by the assumption of non-increasing of sequence $(p_1, p_2, \ldots, p_n)$ we have $i < j$. By the definition of function $T$, variable $r$ is non-decreasing with time, therefore we have $d_i \leq d_j$.

Consider the case $A$ and $B$ are not necessarily singletons. If $p(A) > p(B)$, we have to show that there is a world $w \in A$ such that for all $v \in B$ $d(w) \leq d(v)$. Suppose the contrary, there is a world $v \in B$ such that for all $w \in A$, $d(v) < d(w)$. Let $d(v) = r$. By hypothesis $A \subseteq \{\omega | d(\omega) \geq r + 1\}$. Therefore,

$$p(\{\omega | d(\omega) \geq r + 1\}) \geq p(A) \geq p(B) \geq p(v). \quad (2)$$

But we have $d(v) = r$. Let $M^{(i)}$ denotes the value of variable $M$ at step $i$. We can easily prove

$$M^{(i)} = \sum_{j=i+1}^{n} p_j. \quad (3)$$

If $d(\omega_i) = r$ and $d(\omega_{i+1}) = r + 1$ then because $(p_i)$ is a non-decreasing sequence, we have

$$p(v) \geq p(\omega_i). \quad (4)$$

And by the definition of function $T$,

$$p(\omega_i) > M^{(i)} = \sum_{j=i+1}^{n} p_j = p(\{\omega|d(\omega) \geq r+1\}) \quad (5)$$

But this equation in combination with (2) contradicts equation (4).

*Least coarse.* It is straightforward to show that $|T(p)| = |L_p| + 1$. By the lemma, we have for all congruent transformations $T'$, $|T'(p)| \leq |L_p| + 1$. That means $T$ defined is least coarse. ∎

**Example:** For $\Omega = \{\omega_1, \omega_2, \omega_3, \omega_4\}$

| $i$ | $p_i$ | $M_i$ | $\delta_i$ | $\delta_{\epsilon=.2}$ |
|---|---|---|---|---|
| 1 | 0.5185 | 0.4815 | 0 | 0 |
| 2 | 0.2308 | 0.2507 | 1 | 0 |
| 3 | 0.1538 | 0.0969 | 1 | 1 |
| 4 | 0.0969 | 0 | 2 | 1 |

In this table, the first two and the fifth columns are taken from the example in section 1. The third and fourth columns illustrate how the transformation algorithm works. The difference between the fourth and the fifth columns tells the advatages of $T$ transformation over those by $\epsilon$−rules. The former offers an order-preserving disbelief assignment while the latter does not always guarantee that. When both have such property, the former produces at least as many levels as the latter.

## 3 From Spohnian Disbelief to Probability

In this section, we shall consider the reverse problem, given a Spohnian disbelief function $\delta$, how can we define an "equivalent" probability function. Such transformations are called for if we are to make decision based on the ordinal information provided by disbelief functions and also wish our decision making be immune from attacks of the Dutch book argument. Snow [18] studies a transformation from qualitative probability to ordinary (quantitative) probability for the same reason. In short, we shall consider functions

$$S : \Delta \to \mathcal{P}$$

**Definition 5 (Principle of ordinal congruence II)**
*We say transformation $S$ is congruent if and only if whenever $\delta(A) < \delta(B)$, we have $S(\delta)(A) > S(\delta)(B)$ for all $\delta \in \Delta$ and $A, B \subseteq \Omega$.*

Notice that the strict inequality in $\delta(A) < \delta(B)$ should not be replaced by the weaker inequality $\leq$ as this will have an undesirable consequence. Take for example, two sets $C$ and $D$ such that $C \supset D$ and $\delta(C) = \delta(D)$, the principle of congruence would require $S(\delta)(C) = S(\delta)(D)$. That means $S(\delta)(C - D) = 0$, or in words, probability of all worlds in $(C - D)$ should be 0.

Given a disbelief function $\delta$, let $s = \max \{\delta(\omega)|\omega \in \Omega\}$, i.e., $s$ equals the disbelief value of a least believed world according to $\delta$. We can define $s$ numbers $(k_0, k_1, \ldots, k_s)$, where $k_i = |\{\omega|\delta(\omega) = i\}|$. In other words, we can summarize a Spohnian disbelief function by corresponding vector $(k_0, k_1, \ldots, k_s)$.

Now we want to ponder about the numerical restrictions that the principle of ordinal congruence imposes on probabilities possibly assigned to the worlds in $\Omega$. Let $\Omega_i = \{\omega \in \Omega|\delta(\omega) = i\}$ for $0 \leq i \leq s$ and let $\bar{p}_s$ be the average probability of the worlds in $\Omega_s$. We have $p(\Omega_s) = k_s.\bar{p}_s$ by the addition rule of probability. For any $\omega_{s-1} \in \Omega_{s-1}$, the principle of congruence forces $p(\omega_{s-1}) > p(\Omega_s)$. Therefore, $p(\Omega_{s-1}) > k_{s-1}.k_s.\bar{p}_s$. For $\omega_{s-2} \in \Omega_{s-2}$, $p(\omega_{s-2}) > p(\Omega_{s-1}) + p(\Omega_s)$. Thus, $p(\Omega_{s-2}) > k_{s-2}(k_{s-1} + 1).k_s.\bar{p}_s$. We can continue to write similar inequalities for $p(\Omega_{s-3})$, down to $p(\Omega_0)$. Now we sum those inequalities to get

$$\sum_{i=0}^{s} p(\Omega_i) > (k_0 + 1).(k_1 + 1)\ldots(k_{s-1} + 1).k_s.\bar{p}_s.$$

Taking into account $p(\Omega) = 1$, we infer

$$\bar{p}_s < \frac{1}{(k_0 + 1).(k_1 + 1)\ldots(k_{s-1} + 1).k_s}.$$

Similarly, we have for $0 \leq i \leq s$

$$\bar{p}_i < \frac{1}{(k_0 + 1).(k_1 + 1)\ldots(k_{i-1} + 1).k_i}.$$

where $\bar{p}_i$ denotes the average probability of the worlds in $\Omega_i$.

Given that disbelief function $\delta$ is the only relevant information in the construction of a probability function, the probabilities assigned to the worlds having the same disbelief values must be the same, and hence equal to their average. This leads to the following definition.

**Definition 6 ($S$)** *If $\delta(\omega) = i$ then*

$$S(\delta)(\omega) = \frac{1}{k_0 + 1}.\frac{1}{k_1 + 1}\cdots\frac{1}{k_i + 1}.Z \quad (6)$$

*where Z is the normalization constant.*

From Definition 6, it is evident that probabilities are assigned to the worlds asymmetrically in the sense that the probability value assigned to a world believed with degree $i$ depends only on less disbelieved worlds (with disbelief values less than $i$) and not on those that are more disbelieved.

The second observation is that if $k_0 = k_1 = \ldots k_s = k$ then probability of a world disbelieved with degree $i$ has a very convenient form $p_i = (k+1)^{-i}.Z$. In other words, if we assume an "uniform" distribution of disbelief i.e. $|\Omega_0| = |\Omega_1| = \ldots |\Omega_s| = \frac{|\Omega|}{s}$ then probability and degree of disbelief are related through an exponential (or logarithmic) law.

The third and maybe the most interesting feature of transformation $S$ is its sensitivity to just the ordinal information contained in a disbelief function. It is noted that there is an infinite number of disbelief functions that are equivalent in terms of ranking possible worlds because a disbelief degree can be any natural number. We can identify a "canonical" element of such a class by the following densification operator.

**Definition 7 (Densification)** *Disbelief function $\delta_1$ is called densified version of $\delta_0$ (write $\delta_1 = D(\delta_0)$) if and only if*

*(1) $\delta_1(\omega) \leq \delta_1(\omega')$ whenever $\delta_0(\omega) \leq \delta_0(\omega')$ for all $\omega, \omega' \in \Omega$, and*

*(2) the set $\{\delta_1(\omega)\}$ is a set of consecutive non-negative integers.*

Informally, if we view a disbelief function $\delta$ as an arrangement of the set of possible worlds into strata $\Omega_0$, $\Omega_1$ and so on where $\Omega_i = \{\omega \in \Omega | \delta(\omega) = i\}$ then, what densification does is removing the empty strata and then shifting down the higher ones. Clause (1) ensures that as far as the ordinal order among worlds is concerned, a disbelief function and its densified version are equivalent. In terms of vector $(k_0, k_1, \ldots, k_s)$, a dense disbelief function is one that does not have any $k_i = 0$. It is easy to see that we can not make $S$ change the probability assigned to the worlds in $\Omega$ by inserting a number of zero strata into a disbelief function.

Now we will show that transformation $S$ congruent. First, we need a simple lemma.

**Lemma 2** *Let $p = S(\delta)$ and $G_i = \{\omega | \delta(\omega) > i\}$. If $\delta(\omega) = i$ then $p(\omega) > p(G_i)$.*

**Proof:** For $i = s - 1$, we have $G_{s-1} = \{\omega | \delta(\omega) = s\}$ where For each $\omega \in G_{s-1}$ we have

$$p(\omega) = \frac{1}{k_0 + 1} \cdot \frac{1}{k_1 + 1} \cdots \frac{1}{k_s + 1} . Z$$

Because $|G_{s-1}| = k_s$, we have

$$p(G_{s-1}) = \frac{1}{k_0 + 1} \cdot \frac{1}{k_1 + 1} \cdots \frac{1}{k_{s-1} + 1} \frac{k_s}{k_s + 1} . Z$$

So, if $\delta(\omega) = s - 1$ then $p(\omega) > p(G_{s-1})$.

Similarly, by backward recursion, we can prove for all $i$, if $\delta(\omega) = i$ then $p(\omega) > p(G_i)$. ∎

**Theorem 2** *The $S$ defined above is a congruent Spohnian disbelief to probability transformation function.*

**Proof:** Suppose $\delta(C) < \delta(D)$ where $C$ and $D$ are subsets of $\Omega$. Let $\delta(C) = i$, by axiom S2, there is a $w \in C$ such that $\delta(w) = i$. Since $\delta(D) > i$, for all $v \in D$, $\delta(v) > i$. Let $G_i = \{\omega | \delta(\omega) > i\}$, we have $D \subseteq G_i$. That means $p(G_i) \geq p(D)$ because $p$ is probability function. Since $w \in C$, $p(C) \geq p(w)$. By lemma, $p(w) > p(G_i)$. So, $p(C) > p(D)$. ∎

From the proof of Lemma 2, we can verify that the replacement of a denominator $(k_i + 1)$ in the definition of $S$ with a greater number would leave the transformation congruent. Therefore, we can restore the exponential relationship between probability and disbelief degree, as suggested by Spohn, by choosing one denominator $k_{max}$ where

$$k_{max} = \max_{0 \leq i \leq s} k_i.$$

Thus, another transformation $T'$ can be defined such that for $p' = T'(\delta)$ and $\delta(\omega) = i$,

$$p'(\omega) = \left(\frac{1}{k_{max} + 1}\right)^i . Z \quad (7)$$

So, what is the difference between transformations given by equations (6) and (7)? and what is the rationale for choosing (6) over (7)? One answer is that the former gives less skewed distribution. To see that let us compare the ratios $\frac{p(\omega_1)}{p(\omega_2)}$ and $\frac{p'(\omega_1)}{p'(\omega_2)}$ where $p = S(\delta)$ and $p' = T'(\delta)$ and $\omega_1, \omega_2 \in \Omega$. It is easy to show

**Lemma 3** *If $\delta(\omega_1) > \delta(\omega_2)$ then $\frac{p(\omega_1)}{p(\omega_2)} > \frac{p'(\omega_1)}{p'(\omega_2)}$*

In a sense, transformation $S$ describes a more cautious attitude by allocating to less believed possible worlds the maximal probability that is permitted by the principle of congruence.

Although equation (6) gives a point-valued probability for a possible world disbelieved with degree $i$, in general, for a (non-atomic) proposition disbelieved with degree $i$ what we can infer from the transformation is a range of probabilities. We have the following.

**Corollary 1** *Assuming transformation S, for a disbelief function $\delta$ represented by vector $(k_0, k_1, \ldots, k_s)$, the following statements are equivalent*

(i) $\delta(A) = i$ for $A \subseteq \Omega$;

(ii) $\frac{1}{\prod_{j=0}^{i}(k_j+1)}.Z \leq p(A) < \frac{1}{\prod_{j=0}^{i-1}(k_j+1)}.Z$

**Example:**

| $\delta_i$ | $k_i$ | $p_i$ |
|---|---|---|
| 0 | 1 | $0.5000.Z = 0.5454$ |
| 1 | 2 | $0.1667.Z = 0.1818$ |
| 2 | 1 | $0.0833.Z = 0.0909$ |

where $Z = 0.9167^{-1}$.

While probability to disbelief function transformation is a coarsening process that trades preciseness for computational simplicity, the reverse transformation gives the impression that it imposes unsupported preciseness. It is also evident that $S$ is not the inverse of $T$ i.e. in general we have $S(T(p)) \neq p$. In fact, because $\mathcal{P}$ and $\Delta$ are of different cardinalities, we should not expect to find such a inverse pair of transformations, nevertheless we have $T(S(\delta)) = \delta$ provided $\delta$ is dense.

Because Spohn's definition of deterministic believability (or acceptance) of a proposition by the (set theoretical) inclusion of possible world disbelieved with degree 0, another use of transformation $S$ is in finding the acceptance threshold with the property that the set of higher-than-threshold probability propositions is deductively closed. For example, a probability function transformed from disbelief function with vector $(k_0, k_1, \ldots)$ sets the acceptance threshold at $\frac{k_0}{k_0+1}$. It is worth noting that while the Lottery Paradox establishes that there is no a priori threshold that guarantees deductive closure of the set of higher-than-threshold probability propositions for all probability distributions, given the probability distribution as above, the threshold $\frac{k_0}{k_0+1}$ establishes a cut point of acceptance if we insist that the acceptance relation be closed under logical conjunction.

**Example:**

| $k_0$ | Unnormalized threshold |
|---|---|
| 1 | 0.5000 |
| 2 | 0.6667 |
| 3 | 0.7500 |
| 4 | 0.8000 |

## 4 Transformations and Belief Revision

Any calculus designed to deal with uncertain beliefs has to provide a rule for belief revision due to new information. Such a rule maps an epistemic state to another. Traditionally, for this purpose, Bayes' rule of probability conditioning or its variations are used. But conditioning is not the only rule of belief revision. Lewis [11] proposed an alternative rule called probability *imaging*. For Spohn's disbelief function, conditioning is defined by axiom $S3$.

In the previous sections we proposed some transformations that allows one to shift from one calculus to another, which hopefully, maintains the essence of information. The shifting of calculi creates new problems. Once the same body of information can be represented by different functions and each of them has their own rule for revision in the face of new information, how can we ensure that those revised functions are congruent? In other words, the proposed transformations would be less useful if they did not maintain some kind of consistency with respect to belief revision rules in the different calculi. We will show that they in fact do.

Suppose originally we have disbelief function $\delta$, and after observing evidence (in the form of a proposition) $A$, and applying Spohn's rule of conditioning we get $\delta'$. Further, suppose $\delta$ was transformed to probability function $p$ by $S$. Having observed $A$, we can condition $p$ on the proposition and get $p'$. We will show that $T(p')$ is almost the same as $\delta'$. The result on consistency of the transformation rules $T$ and $S$ can be summarized in the following scheme.

**Theorem 3**

$$T(S(\delta)(.|A)) = D(\delta(.|A))$$

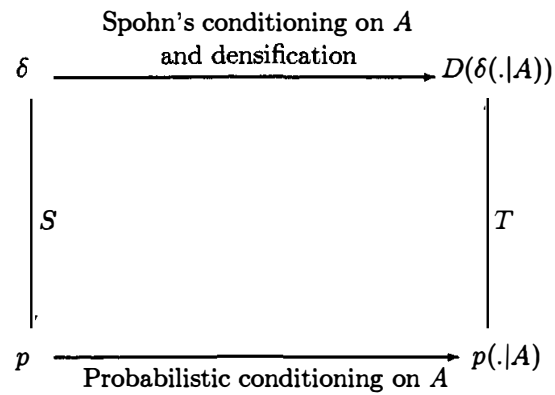

**Proof:** Sketch of proof.

Assume in this proof $\omega, \omega' \in A$. We have: $\delta(\omega|A) \leq \delta(\omega'|A)$ iff $\delta(\omega) \leq \delta(\omega')$ iff $S(\delta)(\omega) \leq S(\delta)(\omega')$ iff $S(\delta)(\omega|A) \leq S(\delta)(\omega'|A)$. The bi-directional inference chain is due by invoking, first, Spohn's conditioning definition, then the definition $S$ and finally, the definition of probabilistic conditioning.

So if $\delta(\omega|A) = \delta(\omega'|A)$ then $S(\delta)(\omega|A) = S(\delta)(\omega'|A)$. And from the later equality, by definition of $T$, we have $T(S(\delta)(\omega|A)) = T(S(\delta)(\omega'|A))$.

Now, to complete the proof we have to show that if $\delta(\omega|A) < \delta(\omega'|A)$ then $T(S(\delta)(\omega|A)) < T(S(\delta)(\omega'|A))$.

Let $\delta(\omega) = i$, so $\delta(\omega') > i$. Recall notation $G_i = \{w \in \Omega | \delta(w) > i\}$. Let $S(\delta) = p$, by lemma 2, we have $p(\omega) > p(G_i)$. We can rewrite

$$p(G_i) = \sum_{\delta(w) > i} p(w).$$

Note that for those $w \in G_i \cap A$, $p(w|A) = \frac{p(w)}{p(A)}$ and for $w \notin G_i \cap A$, $p(w|A) = 0$. So from $p(\omega) > p(G_i)$, we have $\frac{p(\omega)}{p(A)} > \frac{p(G_i \cap A)}{p(A)}$ or using notation of conditional probability $p(\omega|A) > p(G_i \cap A|A)$. This is the condition to increase rank parameter in definition of $T$. So, applying transformation $T$ for probability function $p(.|A)$, we shall have $T(p(.|A)(\omega) < T(p(.|A)(x)$ for all $x \in G_i \cap A$. Because of assumptions $\delta(\omega') > i$ and $\omega' \in A$, we have $\omega' \in G_i \cap A$. Thus, $T(p(.|A)(\omega) < T(p(.|A)(\omega')$. ∎

Similarly, we can prove a result similar to Theorem 3 for the alternative method of probability updating, namely, Lewis' imaging. Lewis [11] motivated his formulation of probabilistic imaging from the fact that conditional probabilities are not the same as probabilities of conditionals except for trivial languages. To define an image of a probability function $p$ on a proposition $A$, one needs, in addition, a *closeness* relation among possible worlds. After observing $A$, the mass (probability) that $p$ assigns to a world $\omega \notin A$ is moved to the world(s) closest to $\omega$ that is(are) in $A$. A disbelief function readily provides a closeness relation. We can define the "distance" between two worlds $\omega_1, \omega_2$ by $|\delta(\omega_1) - \delta(\omega_2)|$. That means, having observed $A$, the mass of a world excluded by that observation will distributed evenly to the remained worlds of its class. Then Theorem 3 stills holds if $(.|A)$ is understood as imaging on $A$ instead of conditioning on $A$. Proof of that fact is similar to the proof of Theorem 3.

## 5 Summary and Conclusion

In this paper we describe two transformations from probability to Spohnian disbelief function and vice-versa. In a departure from the widely used idea of relating disbelief values to infinitesimal probabilities i.e., $\epsilon$-rule, we adopt the principle of ordinal congruence as the basis for the transformations. Transformation $T$ from probability to Spohnian disbelief is obtained if we couple the principle of congruence with the principle of minimal information loss. Transformation $S$ from disbelief to probability function satisfies, in addition to the principle of congruence, the cautious attitude by allocating to less believed possible worlds the maximal probability that is permitted by the principle of congruence. We show that this pair of transformations are consistent with respect to conditionalization.

In the experimental works using $\epsilon$-rule as probability-Spohnian disbelief transformations, a tension is how small should $\epsilon$ be. On one hand, the results from [19, 20, 7] say that with an infinitesimal $\epsilon$, reasonings with probability and Spohnian disbelief are congruent. On the other hand, with $\epsilon$ approaching 0, the transformation defined by $\epsilon$-rule becomes the trivial, i.e., any strictly positive probability is assigned disbelief degree zero. The experiments in [2, 8] are set up, partly, to examine the effect of $\epsilon$ values on reasoning outcome. Unfortunately, the results of those experiments can not provide an unambiguous answer to that tension. In this paper, considering a transformation class broader than that defined by $\epsilon$-rule, we show that $T$ is the best answer.

The transformation from non-probabilistic calculi to probability helps to connect the theories that lack decision methods with the rich body of Bayesian decision theory developed for probability.

Another benefit of such a linkage is that it provides semantics for the values of disbelief functions. For probability, we have the semantics of relative frequency and betting rates. A subjective interpretation of the statement "probability of $A$ is $p$" is the maximum price a risk neutral person would be willing pay to buy a lottery that pays $1.00 if $A$ happens and nothing otherwise. We do not know of any similar semantics for possibility values or Spohnian disbelief values. Indeed, Dubois and Prade [4, 5] take pains to explain that the information content of the numerical value of a possibility function is nothing but an ordinal ranking. Spohn [20] provides a link between his disbelief function and probability. But since this link is not formally stated, it cannot provide a useful semantic for values of disbelief functions via probabilistic semantics.

An advantage of Spohn's epistemic belief theory is computational simplicity. The human mind may not be designed for probabilistic computation. Common sense reasoning may not be probabilistic in nature. These assertions are supported by a large number of empirical studies in decision making [10]. Spohn's the-ory was motivated partly from the notion of representing plain belief. On the other hand, a vast body of literature on the principle of maximization of expected utility suggested by von Neumann and Morgenstern [21] and Savage [12] tend to support the thesis that rational behavior is based on numerical probabilities. So the transformations described in this paper offer a bridge between having plain beliefs and behaving rationally. It also hints at the costs of rational behavior if one starts from plain beliefs.